\documentclass[lettersize,journal]{IEEEtran}
\usepackage{amsmath,amsfonts}
\usepackage{algorithmic}
\usepackage{algorithm}
\usepackage{array}
\usepackage[caption=false,font=normalsize,labelfont=sf,textfont=sf]{subfig}
\usepackage{textcomp}
\usepackage{stfloats}
\usepackage{url}
\usepackage{booktabs} 
\usepackage{tabularx}

\usepackage{verbatim}
\usepackage{graphicx}
\usepackage{cite}
\begin{document}

\title{DWMP: Leveraging Dual World Models for Humanoid Obstacle Traversal}

\author{Rongjun Jin, Jianming Ma, Yue Gao


}

\markboth{Journal of \LaTeX\ Class Files,~Vol.~14, No.~8, July~2026}%
{Shell \MakeLowercase{\textit{et al.}}: A Sample Article Using IEEEtran.cls for IEEE Journals}


\maketitle

\begin{abstract}
Humanoid robots must traverse cluttered obstacle fields using onboard proprioceptive and visual observations, yet existing methods usually process multimodal observations without explicitly considering their different characteristics: proprioceptive observations are low-dimensional but governed by highly nonlinear robot dynamics, while egocentric visual observations are high-dimensional, noisy, and redundant. We propose DWMP (Dual World Model Policy), a framework that provides the actor with separate but complementary world-model representations for humanoid obstacle traversal. A Koopman-based dynamics world model lifts proprioceptive observations into a latent space where their temporal evolution is approximately linear, making the dynamics features easier for the actor to learn from. An RSSM-based visual world model compresses egocentric depth observations into compact stochastic states while preserving obstacle-related geometry. The student policy receives the fused latent representation for action generation, combining linearized proprioceptive dynamics with compressed visual perception. Experiments in simulation and on a Unitree G1 humanoid robot show that DWMP improves obstacle traversal performance over baselines and supports real-world deployment under randomized obstacle layouts.

\end{abstract}


\section{Introduction}
\IEEEPARstart{H}{umanoid} robots are expected to traverse cluttered obstacle fields using only onboard sensing~\cite{rudin2022learning,wu2023robustagile,li2024aicpg,he2024perceptive,gu2023humanoid,haarnoja2024soccer,hoeller2024anymalparkour}. This task requires a policy to jointly reason about two heterogeneous information sources: proprioceptive observations describe the robot's own motion state, while egocentric visual observations provide obstacle geometry. Effective obstacle traversal therefore depends not only on policy learning, but also on representations that match the different characteristics of these two modalities.

Teacher-student distillation is a common route for learning deployable legged-locomotion policies~\cite{wang2024cts,chen2021sim,kumar2021rma,miki2022learning}. A privileged teacher is trained with access to ground-truth terrain or obstacle information, and a student then imitates the teacher using onboard observations. However, most existing humanoid obstacle traversal policies do not explicitly process proprioception and vision according to their different properties. They either compress visual observations while feeding raw proprioception to the actor, or concatenate multimodal features without predictive representation learning. Consequently, the student policy tends to remain a reactive action imitator, instead of learning structured features that describe how the robot state and perceived obstacle layout evolve.

For obstacle traversal, this limitation is particularly restrictive. Proprioceptive observations are low-dimensional, but the underlying robot dynamics are highly nonlinear and difficult for a standard actor network to learn directly from raw inputs~\cite{lusch2018deep,shi2022koopman}. Lifting proprioceptive observations into a latent space with approximately linear temporal evolution can provide the actor with simpler and more predictable dynamics features. In contrast, egocentric depth observations are high-dimensional, noisy, and redundant~\cite{schwarzer2021spr,seo2023masked,micheli2023transformers,hansen2024tdmpc2,brohan2023rt1,brohan2023rt2,openx2024}; they require compression that removes task-irrelevant information while preserving obstacle-related geometry.

We propose DWMP (Dual World Model Policy), a framework that provides the student actor with two complementary world-model representations. A Koopman-based dynamics world model linearizes proprioceptive dynamics in a lifted latent space, while an RSSM-based visual world model compresses depth observations into compact stochastic states and predicts future visual observations. The fused latent representation allows action generation to exploit both predictable body dynamics and compact obstacle-aware perception.

DWMP makes use of the interaction data already produced by the teacher-student pipeline. Trajectories from the teacher's exploratory training phase are used to pre-train the dual world model, and the encoders are further adapted during student policy learning to match the student's own observation distribution. In simulation, DWMP improves obstacle traversal success by about 10\% over strong teacher-student baselines and produces better visual reconstruction on unseen terrains. Real-world tests on a Unitree G1 humanoid robot further show successful traversal on randomized Ceil, Mceilbar, and Narrow obstacle layouts.

The main contributions are: (1) the implementation and real-world deployment, for humanoid obstacle traversal, of a dual world model architecture that models proprioceptive dynamics and visual observations separately; (2) a two-stage training strategy that pre-trains the world models on teacher exploration data before fine-tuning them during student learning, enabling the visual world model to learn more effective representations of obstacle terrain; and (3) simulation and real-world evidence that the resulting representations improve obstacle traversal performance and transfer to randomized physical layouts.
\section{Related works}
\label{sec:related_work}

\subsection{World Models for Robot Dynamics and Perception}

Predictive world models are widely used in model-based reinforcement learning to represent either robot dynamics or high-dimensional observations~\cite{hansen2022tdmpc,seo2023masked,micheli2023transformers,hansen2024tdmpc2,janner2022diffuser}. For dynamics modeling, Koopman operator theory represents nonlinear systems as linear evolution in a lifted observable space~\cite{lusch2018deep}. Deep Koopman models have been applied to robot prediction and control~\cite{shi2022koopman,folkestad2020koopman,bruder2021koopman}, where the linear latent transition is attractive for real-time locomotion control. Recent work further incorporates spectrally constrained Koopman latent dynamics into Dreamer-style world models to improve the stability of long-horizon imagination~\cite{li2026koopmandreamer}. However, these works are usually state-centric and do not address visually guided obstacle traversal.

In parallel, Dreamer-style latent world models learn compact recurrent representations from visual observations~\cite{hafner2020dream,hafner2021mastering,hafner2023mastering,wu2022daydreamer,mendonca2021discovering}. Recent visual representation methods further improve pixel- or depth-based prediction through masked modeling and efficient latent dynamics~\cite{seo2023masked,micheli2023transformers,hansen2024tdmpc2}. Ego-vision world models have also begun to appear in humanoid planning~\cite{liu2025egovision}. These methods demonstrate the value of predictive visual compression, but generic visual world models often mix proprioceptive and exteroceptive information in a single latent state. DWMP instead uses different model structures for the two modalities: Koopman linearization for proprioceptive dynamics and RSSM compression for depth perception.

\subsection{Obstacle Traversal Policies for Humanoid Robots}

Humanoid obstacle traversal requires perception-aware locomotion under partial observability. Deep reinforcement learning has enabled agile legged locomotion and perceptive traversal on uneven terrain~\cite{rudin2022learning,margolis2022walk,siekmann2021sim,wu2023robustagile,li2024aicpg,bellegarda2024visualcpg,haarnoja2024soccer}, while navigation and terrain-perception systems provide useful geometric priors for legged robots~\cite{deluca2023autonomous,chen2024terrainvision,yao2024tail,zhu2025vrrobo}. For deployable policies, teacher-student distillation remains widely used~\cite{wang2024cts,chen2021sim,kumar2021rma,miki2022learning}: a privileged teacher is trained with full state or terrain information, and a student imitates it using onboard observations.

Recent humanoid works follow this paradigm for perceptive locomotion and obstacle-rich environments~\cite{he2024perceptive,gu2023humanoid}. HumanoidPF~\cite{xue2026collision} encodes humanoid-obstacle spatial relationships as collision-free motion directions, and VB-Com~\cite{ren2025vbcom} improves robustness by switching between vision and blind policies under deficient perception. Other methods directly train vision-based policies or imitate human demonstrations~\cite{peng2020ase,escontrela2022learning}. Although these approaches show strong agility, they usually do not explicitly model the transition dynamics of both the robot and the visual scene. DWMP addresses this gap by embedding predictive proprioceptive and visual representations into the deployable student policy.

\section{Method}
\begin{figure*}[t]
{ \centering  
  \includegraphics[width=0.9\linewidth]{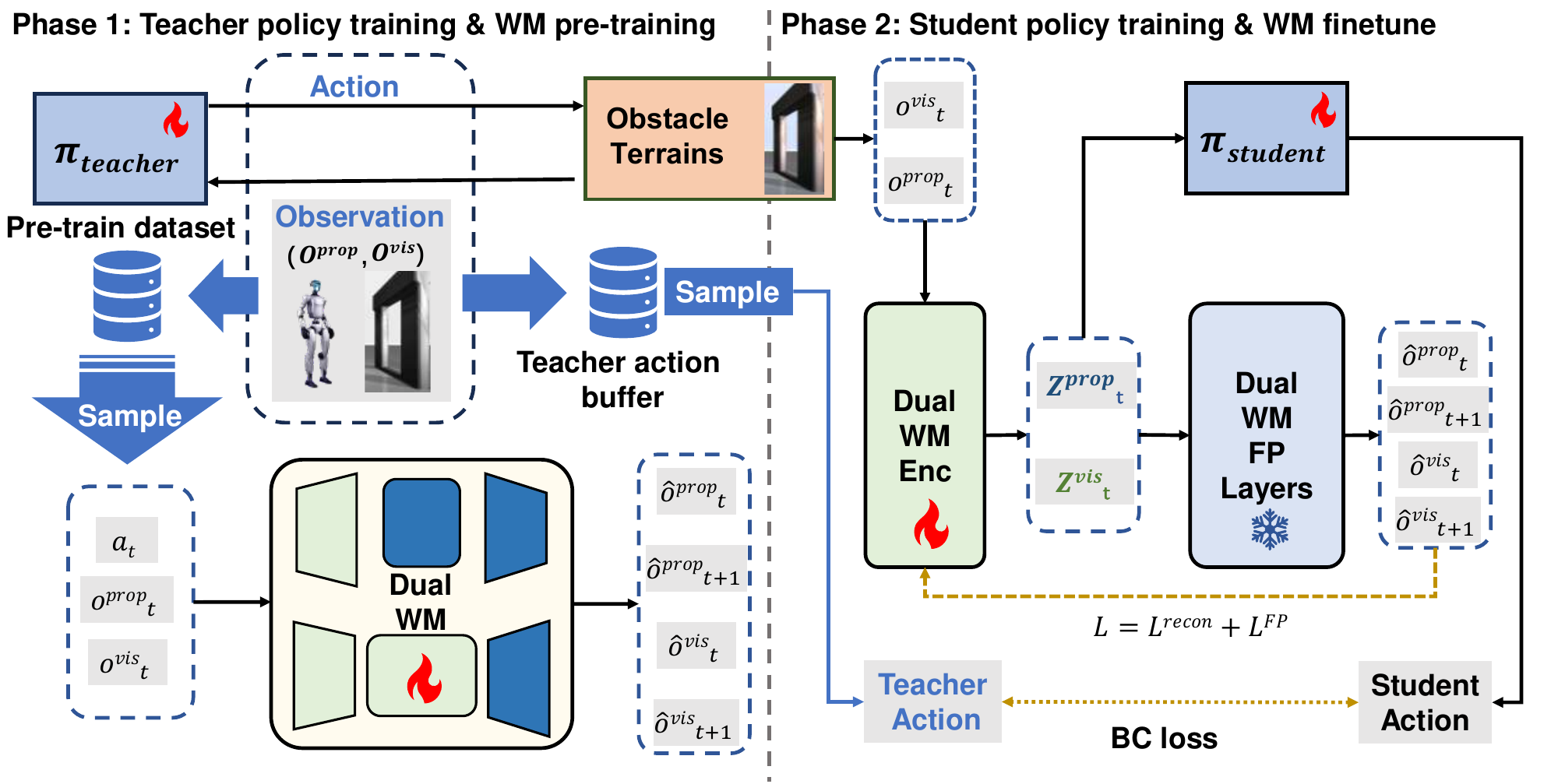}
  \caption{Training pipeline of DWMP. Phase 1: the teacher policy is provided with privileged observations during training, and its interaction data are collected for dual world model pre-training. A trajectory is sampled for the dual world model to reconstruct and predict. Phase 2: the student policy takes non-privileged observations as input. The inputs are encoded by the dual world model network before being passed to the student policy. These encoders remain trainable to adapt to the student's observation distribution, while the forward propagation layers are frozen. }
  \label{fig_Method}
}
\end{figure*}

\subsection{Preliminaries}

\label{sec:preliminaries}


We formulate the humanoid obstacle traversal task as a partially observable Markov decision process (POMDP) defined by the tuple $(\mathcal{O}, \mathcal{A}, \mathcal{T}, \mathcal{R}, \gamma)$.
The observation space $\mathcal{O}$, action space $\mathcal{A}$, and task objective are specified below.

\textbf{Observation space.}
At each timestep $t$, the robot receives a multimodal observation consisting of proprioceptive and exteroceptive components:
\begin{itemize}
    \item \textbf{Proprioceptive observation} $\mathbf{o}_t^{\text{prop}} \in \mathbb{R}^{d_p}$: includes joint positions $\mathbf{q}_t$, joint velocities $\dot{\mathbf{q}}_t$, base orientation implicitly represented by the gravity vector expressed in the body frame $\mathbf{g}_t \in \mathbb{R}^3$, base linear and angular velocities $\mathbf{v}_t, \boldsymbol{\omega}_t \in \mathbb{R}^3$, and the target position relative to the robot's base $\mathbf{p}_t^{\text{target}} \in \mathbb{R}^3$.
    \item \textbf{Visual observation} $\mathbf{o}_t^{\text{vis}} \in \mathbb{R}^{H \times W}$: an egocentric depth image captured from a head-mounted camera, providing geometric information about surrounding obstacles.
\end{itemize}
The full observation is denoted as $\mathbf{o}_t = \{\mathbf{o}_t^{\text{prop}}, \mathbf{o}_t^{\text{vis}}\}$.

\textbf{Action space.}
The action $\mathbf{a}_t \in \mathbb{R}^{d_a}$ specifies target joint positions for a set of proportional-derivative (PD) controllers that actuate the robot's lower-body joints:
\begin{equation}
    \mathbf{a}_t = \mathbf{q}_t^{\text{target}} \in \mathbb{R}^{d_a},
    \label{eq:action_space}
\end{equation}
where $d_a$ corresponds to the number of actuated degrees of freedom.
The PD controllers compute torques $\boldsymbol{\tau}_t$ at a higher control frequency to track the desired joint positions:
\begin{equation}
    \boldsymbol{\tau}_t = \mathbf{K}_p (\mathbf{q}_t^{\text{target}} - \mathbf{q}_t) - \mathbf{K}_d \dot{\mathbf{q}}_t,
    \label{eq:pd_control}
\end{equation}
with $\mathbf{K}_p$ and $\mathbf{K}_d$ denoting the proportional and derivative gain matrices, respectively.

\textbf{Task objective.}
The robot is initialized at a starting position in a cluttered environment containing different kinds of $\mathit{M}$ obstacles $\{\mathcal{B}_i\}_{i=1}^{M}$ of varying sizes and geometries.
A target position $\mathbf{p}_t^{\text{target}} \in \mathbb{R}^3$ is sampled from the surrounding environment.
The objective is to learn a policy $\pi(\mathbf{a}_t \mid \mathbf{o}_t)$ that enables the robot to traverse the obstacle field and reach the target while minimizing collisions.

From the above formulation, we can observe that the observation space of the traversal policy comprises two distinct information streams: proprioceptive observations, which capture the robot's internal dynamics and motion state, and visual observations, which provide exteroceptive perception of the surrounding environment.
These two modalities exhibit fundamentally different characteristics: proprioceptive states evolve according to well-defined physical dynamics that can be compactly represented in a lifted linear space, while visual observations undergo complex appearance changes governed by the robot's ego-motion and the three-dimensional scene geometry.

\subsection{Koopman based Dynamic world model}
For the nonlinear dynamics underlying proprioceptive observations, conventional linear neural networks often struggle to capture the complex transition patterns. To address this limitation, Koopman operator theory proposes a principled approach that lifts the nonlinear system into a higher-dimensional space where the dynamics evolve linearly~\cite{lusch2018deep,shi2022koopman}, thereby improving the capacity of neural networks to model the underlying dynamical system.
\subsubsection{Koopman Operator Theory}
For a nonlinear dynamical system defined by the state transition:
\begin{equation}
    \mathbf{o}^{prop}_{t+1} = \mathbf{F}(\mathbf{o}^{prop}_t),
    \label{eq:nonlinear_dyn}
\end{equation}
where $\mathbf{o}^{prop}_t \in \mathcal{S} \subset \mathbb{R}^{n}$ is the system state at time $t$, and $\mathbf{F}: \mathcal{S} \to \mathcal{S}$ is a nonlinear transition function.
The Koopman operator $\mathcal{K}$ provides an alternative perspective by lifting the dynamics to an infinite-dimensional space of observable functions.
Formally, let $g: \mathcal{S} \to \mathbb{R}$ be a scalar observable function. The Koopman operator acts on $g$ as:
\begin{equation}
    (\mathcal{K} g)(\mathbf{o}^{prop}_t) = g(\mathbf{F}(\mathbf{o}^{prop}_t)) = g(\mathbf{o}^{prop}_{t+1}),
    \label{eq:koopman_def}
\end{equation}
which advances the observable $g$ forward in time.
Crucially, $\mathcal{K}$ is a \emph{linear} operator even when the underlying dynamics $\mathbf{F}$ are nonlinear, enabling linear analysis and prediction in the lifted space.

\subsubsection{Auto-Koopman network}
In practice, we approximate the infinite-dimensional Koopman operator with a finite-dimensional representation using deep learning~\cite{lusch2018deep,shi2022koopman}.
As shown in Fig.~\ref{fig_network}, the Auto-koopman dynamic world model network is composed by an encoder $\phi: \mathcal{S} \to \mathbb{R}^{d_k}$ that maps the original state to a $d_k$-dimensional latent space, and a linear transition matrix $\mathbf{K} \in \mathbb{R}^{d_k \times d_k}$ such that:
\begin{equation}
    \phi(\mathbf{o}^{prop}_{t+1}) \approx \mathbf{K} \, \phi(\mathbf{o}^{prop}_t).
    \label{eq:koopman_approx}
\end{equation}
The first layer of encoder is an observation function transfer layer (OFTL), which transforms each input dimension into its corresponding observation function values. For an input dimension \(i\) with value \(x_i\), this layer outputs, for example,
\begin{equation}
    OFTL(x_i) = \begin{bmatrix} \cos(x_i) \\ \sin(x_i) \\ e^{x_i} \end{bmatrix}.
\end{equation}

The original state can be recovered through a decoder $\psi: \mathbb{R}^{d_k} \to \mathcal{S}$. Training objective of the network is:
\begin{equation}
    \begin{aligned}
    \mathcal{L}_{\text{koopman}} = \mathbb{E} \Big[ 
        & \|\mathbf{o}^{prop}_t - \psi(\phi(\mathbf{o}^{prop}_t))\|^2 \\
        & + \|\phi(\mathbf{o}^{prop}_{t+1}) - \mathbf{K}\,\phi(\mathbf{o}^{prop}_t)\|^2 \Big].
    \end{aligned}
    \label{eq:koopman_loss}
\end{equation}
This formulation enables effective representation learning and long-horizon prediction of the robot's proprioceptive states through repeated application of the linear matrix $\mathbf{K}$.

\subsection{RSSM based visual world model}
\subsubsection{Recurrent State-Space Model}

Representative visual world model work like DreamerV3~\cite{hafner2020dream,hafner2023mastering,hansen2024tdmpc2} learns a world model from high-dimensional observations based on a Recurrent State-Space Model (RSSM) framework.
The RSSM maintains a deterministic recurrent state $\mathbf{z}^{\text{vis}}_t$ that summarizes the history of past observations and actions, and a stochastic state $\mathbf{s}_t$ that captures the instantaneous latent representation.
Given a visual observation $\mathbf{o}^{\text{vis}}_t$ and an action $\mathbf{a}_t$, the RSSM unrolls as follows:
\begin{align}
    &\text{Recurrent state:} \quad \mathbf{z}^{\text{vis}}_t = f_{\text{rec}}(\mathbf{s}_{t-1}, \mathbf{z}^{\text{vis}}_{t-1}, \mathbf{a}_{t-1}), \label{eq:rssm_rec} \\
    &\text{Prior:} \quad p(\mathbf{s}_t \mid \mathbf{z}^{\text{vis}}_t) = \mathcal{N}\big(\mu(\mathbf{z}^{\text{vis}}_t), \sigma(\mathbf{z}^{\text{vis}}_t)\big), \label{eq:rssm_prior} \\
    &\text{Posterior:} \quad q(\mathbf{s}_t \mid \mathbf{z}^{\text{vis}}_t, \mathbf{o}^{\text{vis}}_t) = \mathcal{N}\big(\mu(\mathbf{z}^{\text{vis}}_t, \mathbf{o}^{\text{vis}}_t), \sigma(\mathbf{z}^{\text{vis}}_t, \mathbf{o}^{\text{vis}}_t)\big), \label{eq:rssm_post} \\
    &\text{Observation decoder:} \quad \hat{\mathbf{o}}^{\text{vis}}_t = f_{\text{dec}}(\mathbf{z}^{\text{vis}}_t, \mathbf{s}_t), \label{eq:rssm_dec} \\
    &\text{Reward predictor:} \quad \hat{r}_t = f_{\text{rew}}(\mathbf{z}^{\text{vis}}_t, \mathbf{s}_t). \label{eq:rssm_rew}
\end{align}
Here, $f_{\text{rec}}$ is a Gated Recurrent Unit (GRU) that updates the deterministic recurrent state, while the prior $p(\mathbf{s}_t \mid \mathbf{z}^{\text{vis}}_t)$ and posterior $q(\mathbf{s}_t \mid \mathbf{z}^{\text{vis}}_t, \mathbf{o}^{\text{vis}}_t)$ are diagonal Gaussian distributions parameterized by neural networks.
During training, the posterior incorporates the current observation $\mathbf{o}^{\text{vis}}_t$ to produce a more informed stochastic state; during imagination, only the prior is available.
The RSSM is trained by minimizing a variational bound consisting of reconstruction loss for $\mathbf{o}^{\text{vis}}_t$, prediction loss for $\hat{r}_t$, and a KL divergence regularizer between the posterior and prior distributions:
\begin{equation}
    \begin{aligned}
    \mathcal{L}_{\text{rssm}} = \mathbb{E} \Big[ 
        &-\ln f_{\text{dec}}(\mathbf{o}^{\text{vis}}_t \mid \mathbf{z}^{\text{vis}}_t, \mathbf{s}_t) 
        -\ln f_{\text{rew}}(r_t \mid \mathbf{z}^{\text{vis}}_t, \mathbf{s}_t) \\
        &+ \beta \, D_{\text{KL}}\big(q(\mathbf{s}_t \mid \mathbf{z}^{\text{vis}}_t, \mathbf{o}^{\text{vis}}_t) \,\|\, p(\mathbf{s}_t \mid \mathbf{z}^{\text{vis}}_t)\big) \Big],
    \end{aligned}
    \label{eq:rssm_loss}
\end{equation}
where $\beta$ is a hyperparameter that controls the capacity of the latent information channel.

\subsubsection{Depth Dreamer network}
\begin{figure}[!t]
{ \centering  
  \includegraphics[width=0.95\linewidth, trim=10 10 10 10, clip]{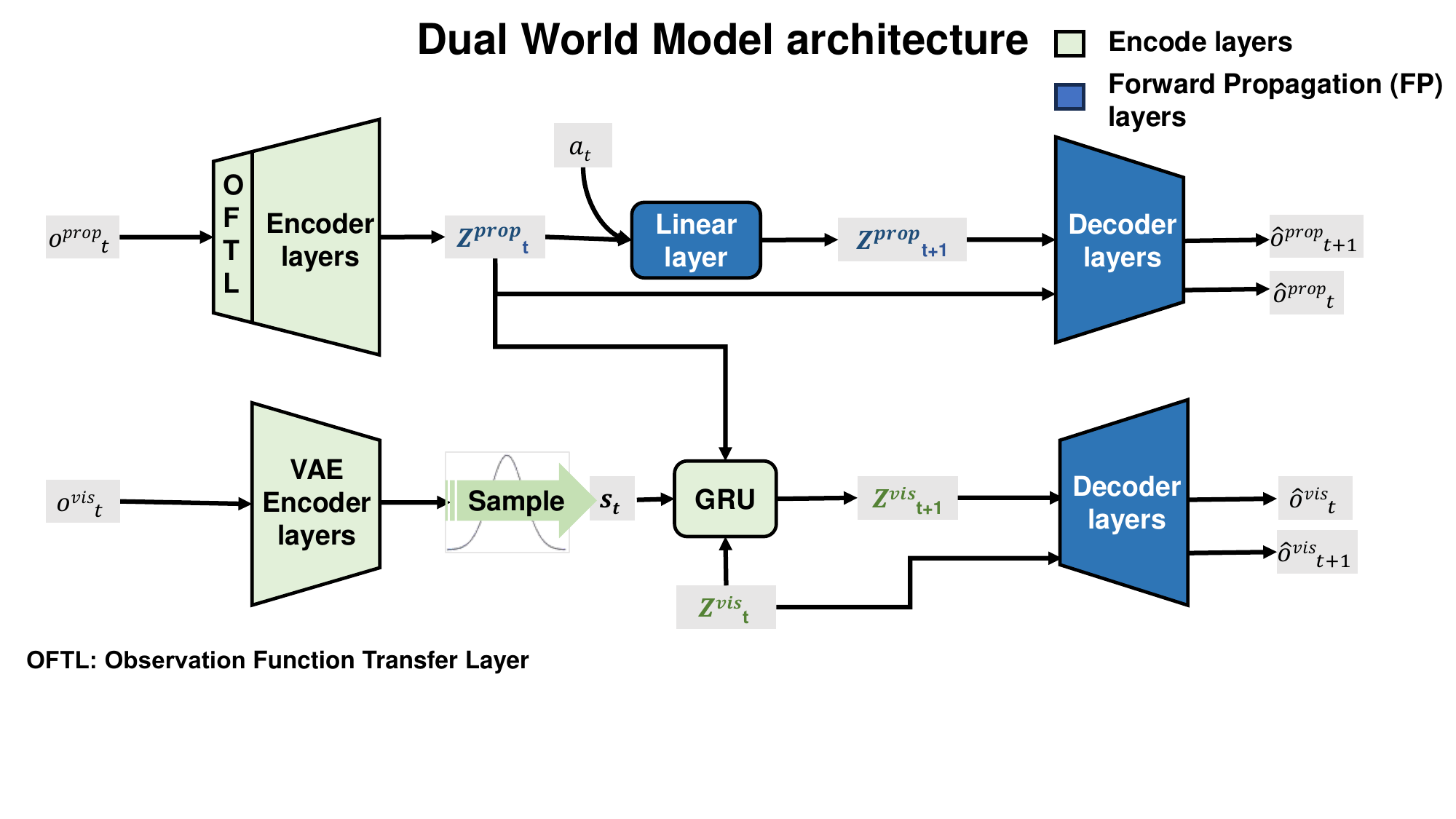}
  \caption{The dual world model network consists of an Auto-Koopman network (on top) and a Depth Dreamer network, both networks are trained for reconstruction and prediction.}
  \label{fig_network}
}
\end{figure}

Depthdreamer is a Dreamer-style depth image representation and prediction network based on the RSSM framework. As shown in Fig.~\ref{fig_network}, the model consists of the following components. A VAE encoder processes the input depth map $\mathbf{o}^{vis}_t$ through several convolutional layers followed by two fully-connected branches that output the mean and log-variance of the posterior distribution $q(\mathbf{s}_t \mid \mathbf{z}^{vis}_t, \mathbf{o}^{vis}_t)$. A GRU updates the deterministic recurrent state $\mathbf{z}^{vis}_{t+1} = f_{\text{rec}}(\mathbf{s}_{t}, \mathbf{z}^{vis}_{t}, \mathbf{a}_{t})$. A prior network (a two-layer MLP) maps $\mathbf{s}_t$ to the parameters of the prior $p(\mathbf{s}_t \mid \mathbf{z}^{vis}_t)$. During training, the stochastic state $\mathbf{s}_t$ is sampled from the posterior via reparameterization; during imagination, it is sampled from the prior. The concatenation $[\mathbf{s}_t, \mathbf{z}^{vis}_t]$ is fed into a decoder (with transposed convolutions) to reconstruct the depth observation $\hat{\mathbf{o}}_t$, and into a small MLP to predict the reward $\hat{r}_t$. The whole network is trained end-to-end by minimizing the loss in Eq.~\eqref{eq:rssm_loss}.

\subsection{Dual World Model for Obstacle Traversal Policy}
\label{sec:dual_world_model}

DWMP uses a dual world model to represent and predict both proprioceptive dynamics and visual information during obstacle traversal. The Koopman-based dynamics world model maps proprioceptive observations into a linear latent space, while the DepthDreamer visual world model compresses high-dimensional depth images into compact stochastic representations. Together, the two models provide the traversal policy with structured and predictive inputs. 

\subsubsection{Policy Input from Dual-World Model Representations}

At each timestep $t$, the robot receives raw observations $\mathbf{o}_t = \{\mathbf{o}_t^{\text{prop}}, \mathbf{o}_t^{\text{vis}}\}$.
Rather than feeding these raw signals directly to the policy, DWMP first processes them through the respective world models to extract predictive representations:

\begin{itemize}
    \item \textbf{Koopman latent state:} The proprioceptive observation $\mathbf{o}_t^{\text{prop}}$ is encoded via the Auto-Koopman network into a lifted latent state $\mathbf{z}^{prop}_t = \phi(\mathbf{o}_t^{\text{prop}}) \in \mathbb{R}^{d_k}$. This representation captures the robot's instantaneous dynamics in a linear, temporally predictable form.
    \item \textbf{RSSM latent state:} The depth image $\mathbf{o}_t^{\text{vis}}$ is encoded through the RSSM posterior network to compute $ \mathbf{z}_t^{\text{vis}}$. This compact representation summarizes the spatial structure of the surrounding obstacles.
\end{itemize}

The full policy input is constructed as:
\begin{equation}
    \mathbf{x}_t^{\text{policy}} = \big[\mathbf{z}_t^{\text{prop}} \; \mathbf{z}_t^{\text{vis}}\big],
    \label{eq:policy_input}
\end{equation}
This fused representation provides the policy with two critical capabilities: anticipating how it will move in the near future via the Koopman latent states, and perceiving what obstacles lie ahead via the compressed depth representation $\mathbf{z}_t^{\text{vis}}$.

The policy network $\pi_\theta$ is implemented as a multi-layer perceptron (MLP) that outputs the parameters of a Gaussian distribution over target joint positions:
\begin{equation}
    \mathbf{a}_t \sim \pi_\theta(\mathbf{a}_t \mid \mathbf{x}_t^{\text{policy}}) = \mathcal{N}\big(\boldsymbol{\mu}_\theta(\mathbf{x}_t^{\text{policy}}),\; \boldsymbol{\sigma}_\theta(\mathbf{x}_t^{\text{policy}})\big),
    \label{eq:policy_output}
\end{equation}
where $\mathbf{a}_t \in \mathbb{R}^{d_a}$ defines the target joint positions for the PD controllers.

\subsubsection{Co-Training Loop}
\label{sec:co_training}

The dual world models and the traversal policy are trained within a unified co-training loop, rather than in separate stages.

The co-training procedure proceeds in three alternating phases, as illustrated in Algorithm~\ref{alg:student_training}:

\textbf{Phase 1: Teacher policy training and WM pre-training. }
A teacher policy $\pi_{\text{teacher}}$ with access to privileged environment information (e.g., ground-truth obstacle positions and heights) is trained using reinforcement learning.
Critically, interaction data are collected $\mathcal{D}_{\text{explore}} = \{(\mathbf{o}_t^{\text{prop}}, \mathbf{o}_t^{\text{vis}}, \mathbf{a}_t)\}$ from the training phase of teacher policy.
This data contains rich exploratory behaviors and diverse state transitions that are essential for building generalizable world models.
The environment resets when the robot collides with obstacles or reaches the target, generating trajectories of varying lengths and outcomes.

Using the collected dataset $\mathcal{D}_{\text{explore}}$, we jointly pre-train the two world models.
The Koopman dynamics model is optimized with the loss defined in Eq.~\eqref{eq:koopman_loss}, learning to reconstruct proprioceptive states and predict their linear evolution.
The DepthDreamer visual model is optimized with the loss defined in Eq.~\eqref{eq:rssm_loss}, learning to reconstruct depth images and predict future latent states conditioned on the Koopman latent state as the action signal:
\begin{equation}
    \mathbf{z}_t^{\text{vis}} = f_{\text{rec}}(\mathbf{z}_{t-1}^{\text{vis}}, VAE(\mathbf{o}_{t}^{\text{vis}}), \mathbf{z}^{prop}_{t-1}).
    \label{eq:vis_rec_conditioned}
\end{equation}
The total world model loss is a weighted sum:
\begin{equation}
    \mathcal{L}_{\text{world}} = \mathcal{L}_{\text{koopman}} + \alpha_{\text{vis}} \, \mathcal{L}_{\text{rssm}},
    \label{eq:world_loss}
\end{equation}
where $\alpha_{\text{vis}}$ balances the two objectives.

\textbf{Phase 2: Student policy training and WM finetuning.}
\begin{algorithm}[t]
\caption{Student Policy Training}
\label{alg:student_training}
\begin{algorithmic}[1]
\REQUIRE Pre-trained world models (Koopman $+$ RSSM), teacher dataset $\mathcal{D}_{\text{teacher}}$
\ENSURE Student policy $\pi_\theta$

\STATE Initialize $\pi_\theta$, replay buffer $\mathcal{D}_{\text{student}} \leftarrow \emptyset$

\FOR{iteration $= 1$ to $N$}
    \STATE Rollout $\pi_\theta$ in environment, $a_t = \pi_{\theta}(\mathbf{x}_t^{\text{policy}})$
     \STATE Store transition $(\mathbf{o}_t^{\text{prop}}, \mathbf{o}_t^{\text{vis}}, \mathbf{a}_t, r_t, \mathbf{o}_{t+1}^{\text{prop}}, \mathbf{o}_{t+1}^{\text{vis}})$ in $\mathcal{D}_{\text{student}}$
    \STATE Update Koopman model and RSSM on $\mathcal{D}_{\text{student}}$

    \STATE Update $\pi_\theta$ with $\mathcal{L}_{\text{student}}$ on $\mathcal{D}_{\text{student}}$ and $\mathcal{D}_{\text{teacher}}$
\ENDFOR

\RETURN $\pi_\theta$
\end{algorithmic}
\end{algorithm}
Once the world models are pre-trained, a student policy $\pi_\theta$ will use the world models for providing fused representations $\mathbf{x}_t^{\text{policy}}$ constructed in Eq.~\eqref{eq:policy_input} and will receive no privileged information.

During student policy training process shown in \ref{alg:student_training}, at each timestep the raw observations $\mathbf{o}_t^{\text{prop}}$ and $\mathbf{o}_t^{\text{vis}}$ are first passed through their respective world model encoders to produce $\mathbf{z}^{prop}_t$ and $\mathbf{z}_t^{\text{vis}}$. The policy then samples an action $\mathbf{a}_t \sim \pi_\theta(\cdot \mid \mathbf{x}_t^{\text{policy}})$, which is executed in the simulator to obtain the next observations and reward.

Simultaneously, as the student policy interacts with the environment, all transition data $\{(\mathbf{o}_t^{\text{prop}}, \mathbf{o}_t^{\text{vis}}, \mathbf{a}_t, \mathbf{o}_{t+1}^{\text{prop}}, \mathbf{o}_{t+1}^{\text{vis}}, r_t)\}$ generated by the student policy are stored in a replay buffer $\mathcal{D}_{\text{student}}$.
The world models are updated on this buffer, allowing them to adapt to the student's specific action distribution and the states it encounters.
This online adaptation is crucial because the student policy, being less capable than the privileged teacher during early training, may visit states that were absent from the teacher's exploratory data.

The student policy $\pi_\theta$ is optimized using a composite objective that combines reward maximization with behavior cloning from the teacher:
\begin{equation}
    \begin{aligned}
    \mathcal{L}_{\text{student}}(\theta) = 
    &-\mathbb{E}_{\pi_\theta} \Big[ \sum_{t=0}^{T} \gamma^t r_t \Big] \\
    &- \lambda_{\text{bc}} \, \mathbb{E}_{\mathcal{D}_{\text{teacher}}} \Big[ \log \pi_\theta(\mathbf{a}_t^{\text{teacher}} \mid \mathbf{x}_t^{\text{policy}}) \Big],
    \end{aligned}
    \label{eq:student_loss}
\end{equation}

where the first term is the cumulative task reward, and the second term is the behavior cloning loss.

\section{Experiments}


\subsection{Experiment Setup}
We employ the Unitree G1 humanoid robot to perform obstacle traversal tasks.
Training is conducted in the MuJoCo physics engine as our simulation environment.
A variety of obstacle terrain blocks are generated within the environment, and in each episode the robot is commanded to reach a specified target position.

\begin{figure}[!t]
{ \centering  
  \includegraphics[width=2.5in]{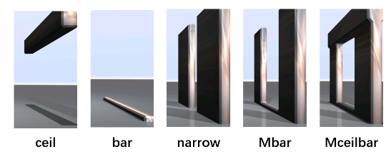}
  \caption{Obstacle types used in obstacle traversal capability experiment.}
  \label{fig_obstacles}
}
\end{figure}

\subsection{Effect of Two-Stage Training on Visual Reconstruction}

To evaluate the effect of the proposed two-stage training strategy, we compare the visual world model during student-stage fine-tuning under two initialization settings: with Phase 1 pre-training on teacher exploration data (warm up), and without such pre-training. For each validation episode, the validation reconstruction loss is computed by averaging the compression-reconstruction loss of the visual world model over all timesteps in that episode.

\begin{figure}[!t]
\centering
\includegraphics[width=\linewidth]{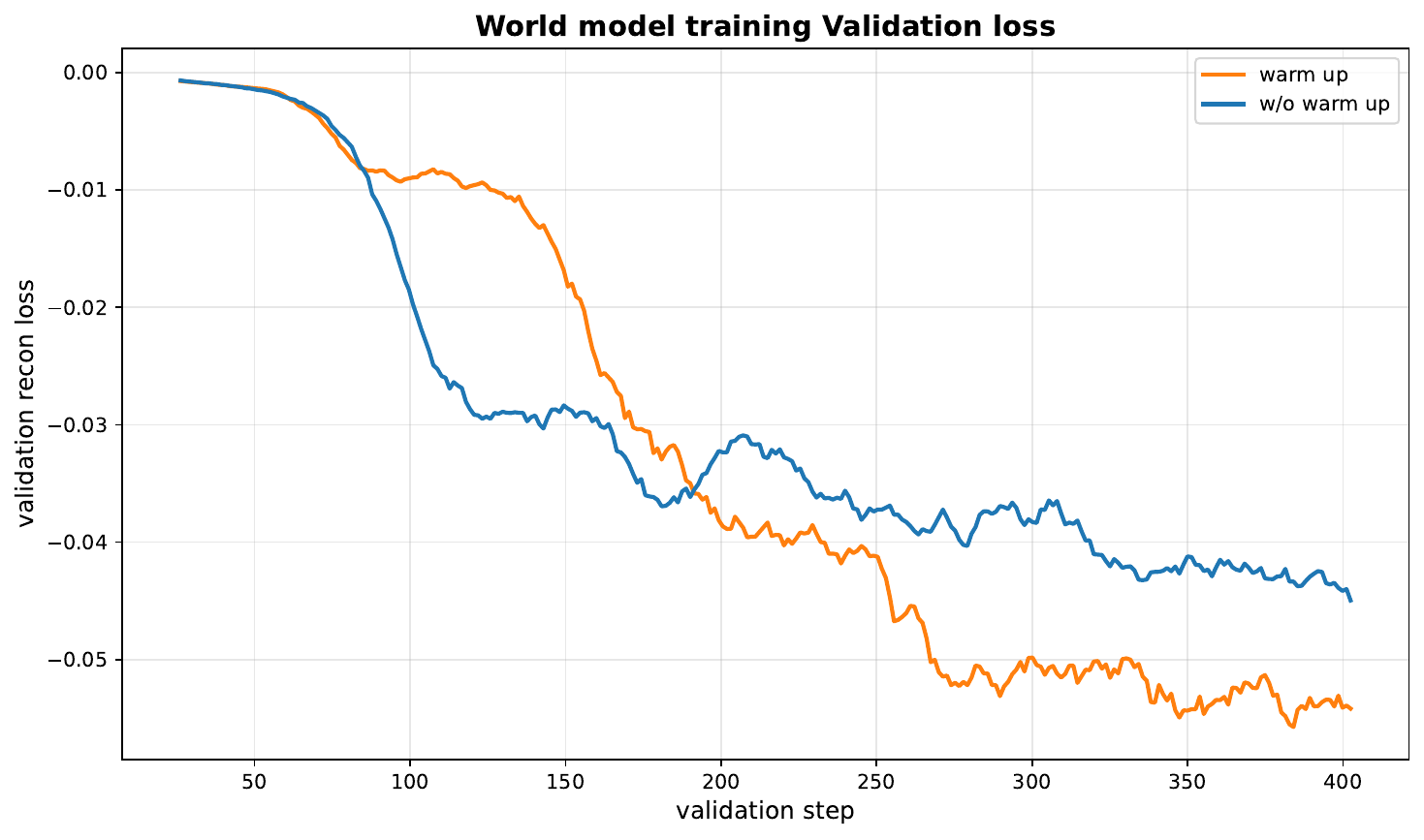}
\caption{Validation reconstruction loss during student-stage fine-tuning with and without Phase 1 warm up. Each validation value is obtained by averaging the visual world model's compression-reconstruction loss over one episode. The warm-up initialization ultimately achieves a lower reconstruction loss than training without warm up.}
\label{fig:world_model_warmup}
\end{figure}

As shown in Fig.~\ref{fig:world_model_warmup}, both settings reduce the validation reconstruction loss during fine-tuning, whereas the model initialized through Phase 1 pre-training reaches a distinctly lower final loss. This result indicates that teacher exploration data provide a more effective initialization for subsequent adaptation to the student policy's observation distribution. Consequently, the two-stage strategy improves the visual world model's reconstruction quality and strengthens its ability to encode obstacle-terrain information for downstream policy learning.

\subsection{Visual Representation Capability Experiment}
Numerous existing architectures can serve as visual world models. We need to identify a model that best meets the following requirements: 
(1) considering computational cost and inference latency, the model should be as lightweight as possible; 
(2) the model must be able to produce good depth-image representations even on unseen terrains; 
(3) during obstacle traversal, accurate visual representations are most critical when the robot encounters obstacles, so the depth reconstruction loss should be particularly low in those situations.
Given these requirements, we evaluate the following model architectures: 
(1) \textbf{Dreamer-V3}, the classic RSSM-based world model, which takes both proprioception and depth images as input, predicts the next depth frame, and simultaneously compresses and reconstructs depth images into latent variables in real time;
(2) \textbf{DWMP}, the proposed method;
(3) \textbf{VAE}, serving as a baseline, which simply compresses and reconstructs images without any temporal modeling, allowing us to compare the performance difference between a general image representation model and a world model in the context of robotic obstacle traversal.

\begin{figure}[!t]
\centering
\includegraphics[width=0.9\linewidth]{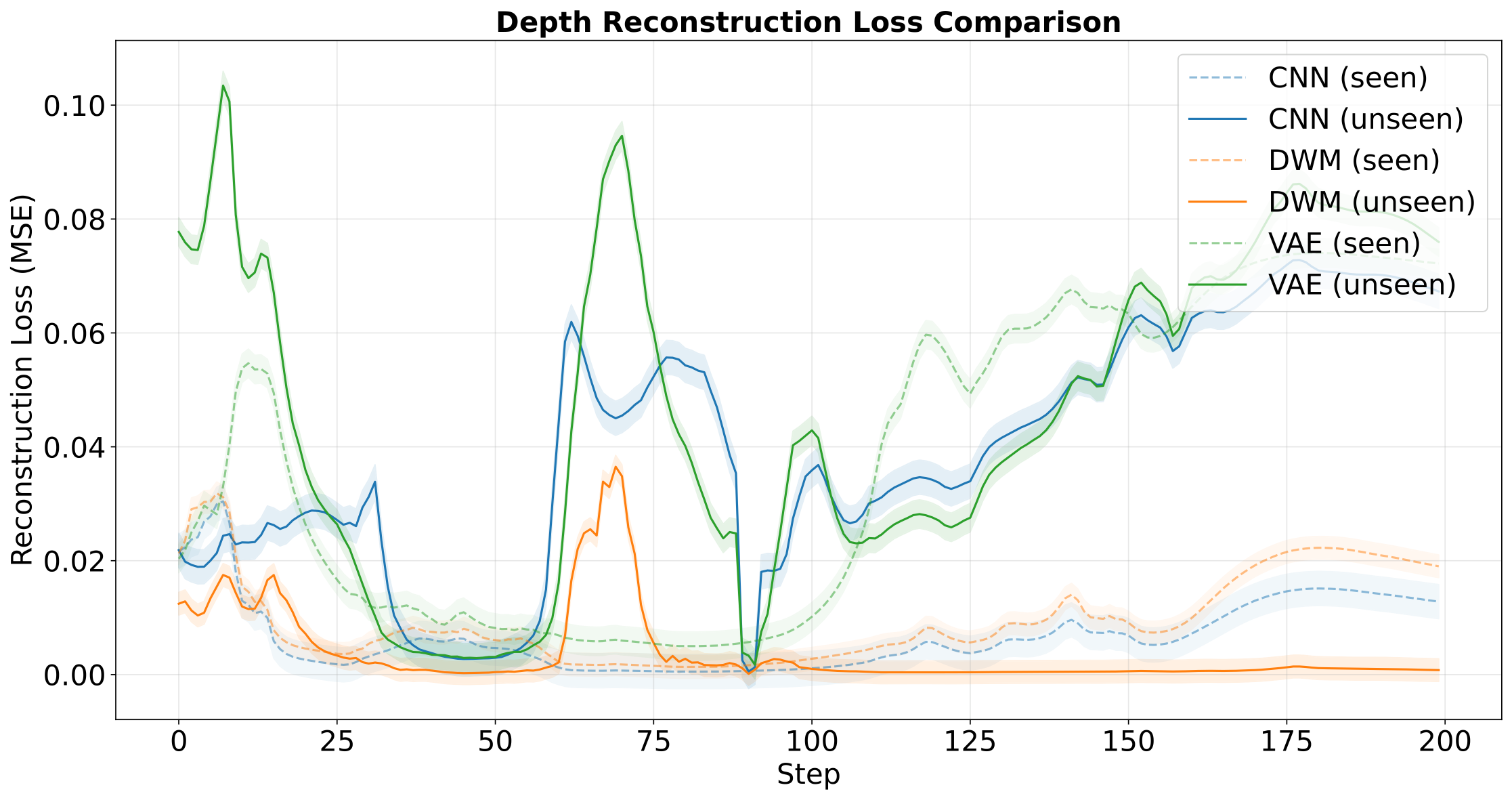}
\caption{Depth reconstruction loss of different visual representation methods on seen and unseen obstacle terrains.}
\label{fig_1}
\end{figure}

As shown in Fig.~\ref{fig_1}, Dreamer-Koopman exhibits better representation capability on unseen obstacle terrains. 
This indicates that during complex locomotion such as obstacle traversal, more effective processing of dynamics information helps the visual perception model understand how the visual scene changes with the robot's motion, thereby providing more valuable visual features to the downstream actor.
Compared with pure image representation models like VAE, both RSSM-based world models achieve better performance, suggesting that learning how the visual observations evolve as the robot moves is beneficial for handling previously unseen visual information in locomotion tasks.
Table~\ref{tab:model_comparison} further compares model inference speed and reconstruction loss under different latent dimensions.

\subsection{Obstacle Traversal Capability Experiment}
We also compare the obstacle traversal capabilities in simulation as reported in Table~\ref{tab:performance_comparison}. 
The results show that, compared with teacher-student distillation methods, our approach achieves a higher success rate on more complex terrains, while also featuring a simpler training procedure.

\begin{table}[!t]
\centering

\caption{Success Rate (SR) of Different Perception World Models in Obstacle Traversal Tasks}
\label{tab:performance_comparison}
\begin{tabular}{lccccc}
\toprule
 & Ceil & Bar & Narrow & Mbar & Mceilbar \\
\midrule
Ours & \textbf{0.95} & 0.85 & \textbf{0.90} & \textbf{0.90} & 0.85 \\
HumanoidPF & 0.90 & 0.85 & 0.85 & 0.80 & \textbf{0.90} \\
Dreamer-v3 & 0.85 & 0.65 & 0.75 & 0.70 & 0.85 \\
VAE & 0.80 & 0.80 & 0.85 & 0.80 & 0.70 \\
Ours (distillation) & 0.85 & \textbf{0.90} & 0.90 & 0.75 & 0.80 \\
\bottomrule
\end{tabular}

\end{table}

\begin{figure}[!t]
{ \centering  
  \includegraphics[width=0.9\linewidth]{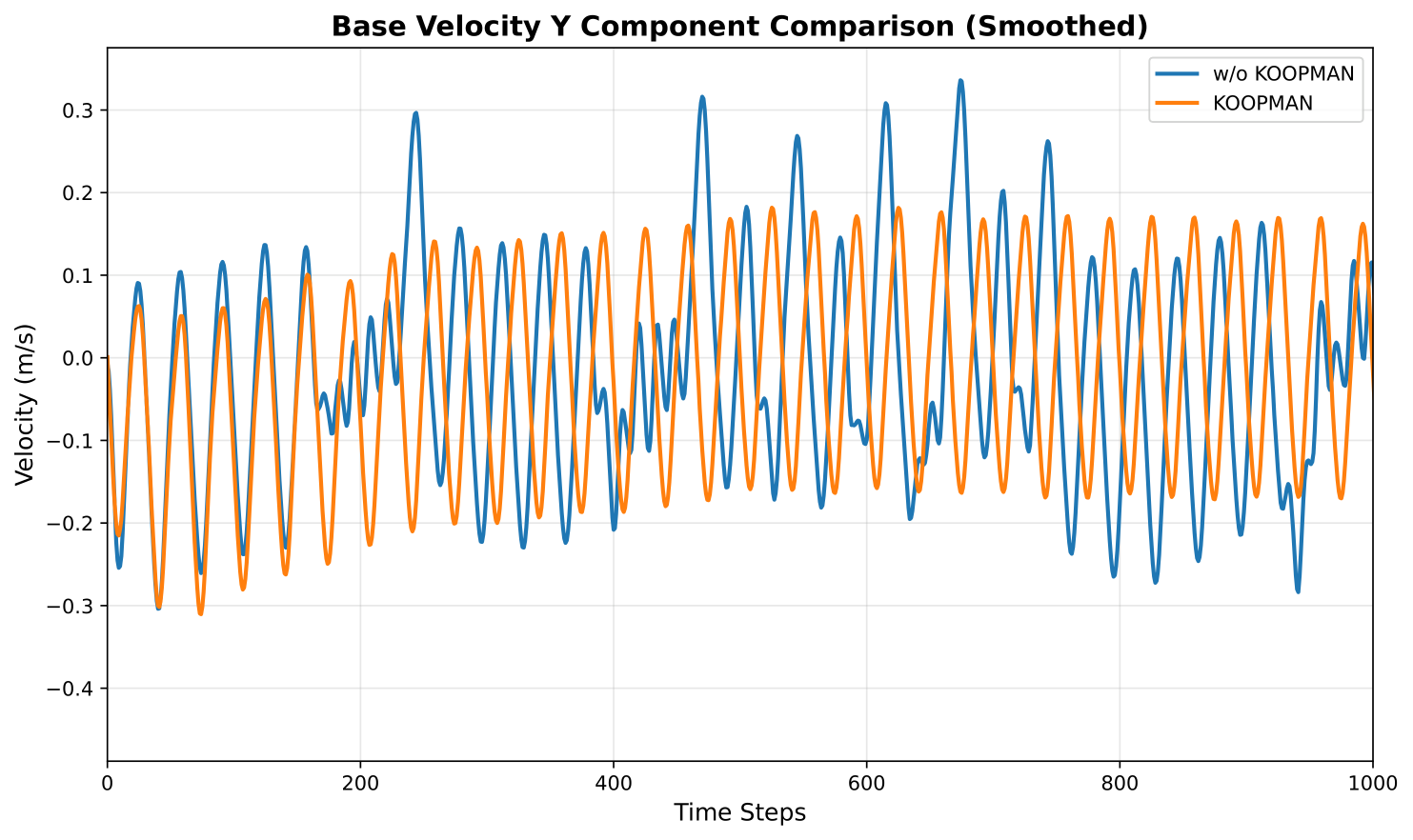}
  \caption{Comparison of Y-axis velocity profiles with and without the Koopman dynamics representation during traversal of the Bar obstacle terrain.}
  \label{fig_2}
}
\end{figure}

\begin{figure}[!t]
\centering
\includegraphics[width=0.9\linewidth, trim=5 5 5 5, clip]{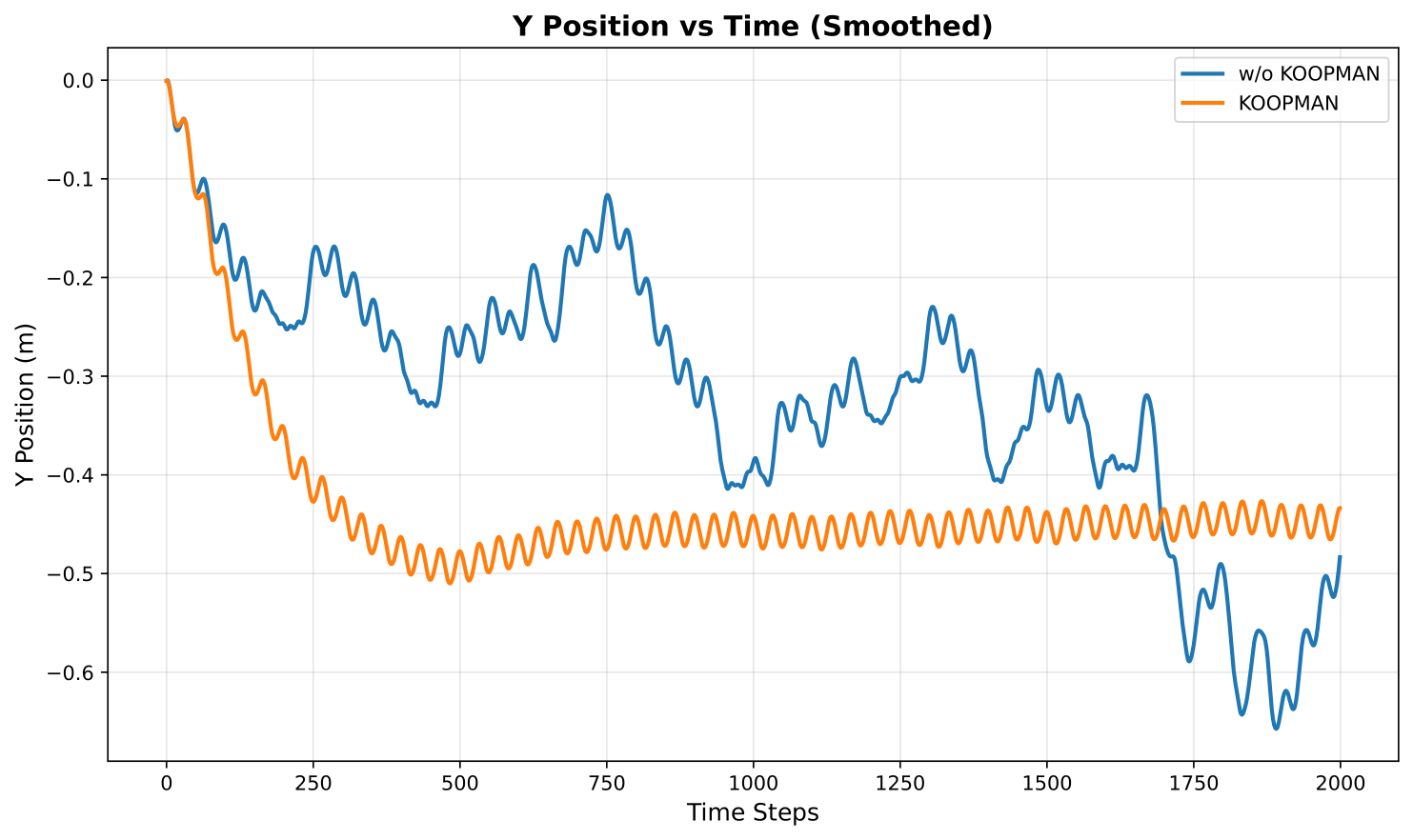}
\caption{Comparison of Y-axis position trajectories with and without the Koopman dynamics representation during traversal of the Bar obstacle terrain.}
\label{fig_3}
\end{figure}

Finally, as shown in Fig.~\ref{fig_2} and Fig.~\ref{fig_3}, the Koopman model provides the actor with more informative dynamics features, resulting in more stable and smooth locomotion performance.

\begin{table}[!t]
\centering
\caption{Inference speed and reconstruction loss of Dreamer-Koopman and DreamerV4.}
\label{tab:model_comparison}
\footnotesize
\setlength{\tabcolsep}{4pt}
\begin{tabular}{@{}lcc@{}}
\toprule
\textbf{Model} & \textbf{Inf. Time (ms)} & \textbf{Max Recon. Loss (MSE)} \\
\midrule
Dreamer-Koopman (2)    & 1.25 & 0.023 \\
Dreamer-Koopman (8)    & 8.67 & 0.018 \\
Dreamer-Koopman (16)    & 22.12 & 0.016 \\
\midrule
DreamerV4 (2)          & 2.38 & 0.027 \\
DreamerV4 (8)          & 24.85 & 0.014 \\
DreamerV4 (16)          & 42.40 & 0.016 \\
\bottomrule
\end{tabular}
\end{table}

\subsection{Real-world Experiment}
We further validate the proposed policy on the Unitree G1 humanoid robot in real-world obstacle traversal experiments. The real-world tests are conducted on three obstacle terrain types, Ceil, Mceilbar, and Narrow. During testing, the obstacle positions are randomly arranged and are not kept identical to the simulation layouts. Despite this layout mismatch, the robot can still use onboard perception to identify the obstacle geometry and traverse the obstacles successfully, indicating that the learned representation is not tied to memorized obstacle placements.

Table~\ref{tab:real_world_performance} reports the pass rate of real-world obstacle traversal. The comparison includes our method and a HumanoidPF direct-distillation baseline, where the deployable policy is trained by directly distilling the privileged HumanoidPF teacher without the proposed dual world model representation.

\begin{table}[!t]
\centering
\caption{Real-world Pass Rate of Obstacle Traversal}
\label{tab:real_world_performance}
\footnotesize
\setlength{\tabcolsep}{4pt}
\begin{tabular}{@{}lccc@{}}
\toprule
\textbf{Method} & \textbf{Ceil} & \textbf{Mceilbar} & \textbf{Narrow} \\
\midrule
DWMP (Ours) & 0.95 & 0.8 & 0.7 \\
HumanoidPF Direct Distill. & 0.8 & 0.6 & 0.65 \\
\bottomrule
\end{tabular}
\end{table}

\section{Conclusion}
This paper presents DWMP, a dual world model framework for humanoid obstacle traversal with onboard proprioceptive and visual observations. DWMP learns structured representations that match the different characteristics of the two observation modalities. By fusing proprioceptive and visual latent representations, the student policy can better exploit both robot dynamics and visual geometry for obstacle traversal. Experiments in simulation show improved traversal success and representation quality over teacher-student baselines, and real-world tests on randomized obstacle layouts further demonstrate the deployability of the proposed framework. Future work will extend the dual world model to more diverse dynamic environments and investigate tighter integration between predictive representations and long-horizon motion planning.

\bibliography{references}

@article{lusch2018deep,
  title={Deep learning for universal linear embeddings of nonlinear dynamics},
  author={Lusch, Bethany and Kutz, J Nathan and Brunton, Steven L},
  journal={Nature Communications},
  year={2018},
  publisher={Nature Publishing Group}
}

@inproceedings{shi2022koopman,
  title={Koopman Operator Based Modeling and Control for Legged Robots},
  author={Shi, Guanya and Meng, Xiangyun and others},
  booktitle={IEEE/RSJ International Conference on Intelligent Robots and Systems (IROS)},
  year={2022}
}

@inproceedings{folkestad2020koopman,
  title={Koopman-based Control of a Soft Robot},
  author={Folkestad, Carl and others},
  booktitle={IEEE International Conference on Robotics and Automation (ICRA)},
  year={2020}
}

@inproceedings{bruder2021koopman,
  title={Koopman-based Model Predictive Control for Legged Locomotion},
  author={Bruder, Daniel and others},
  booktitle={Conference on Robot Learning (CoRL)},
  year={2021}
}

@inproceedings{hafner2020dream,
  title={Dream to Control: Learning Behaviors by Latent Imagination},
  author={Hafner, Danijar and Lillicrap, Timothy and Ba, Jimmy and Norouzi, Mohammad},
  booktitle={International Conference on Learning Representations (ICLR)},
  year={2020}
}

@inproceedings{hafner2021mastering,
  title={Mastering Atari with Discrete World Models},
  author={Hafner, Danijar and Lillicrap, Timothy and Norouzi, Mohammad and Ba, Jimmy},
  booktitle={International Conference on Learning Representations (ICLR)},
  year={2021}
}

@article{hafner2023mastering,
  title={Mastering Diverse Domains through World Models},
  author={Hafner, Danijar and Pasukonis, Jurgis and Ba, Jimmy and Lillicrap, Timothy},
  journal={arXiv preprint arXiv:2301.04104},
  year={2023}
}

@inproceedings{wu2022daydreamer,
  title={DayDreamer: World Models for Physical Robot Learning},
  author={Wu, Philipp and others},
  booktitle={Conference on Robot Learning (CoRL)},
  year={2022}
}

@inproceedings{mendonca2021discovering,
  title={Discovering and Achieving Goals via World Models},
  author={Mendonca, Russell and others},
  booktitle={Advances in Neural Information Processing Systems (NeurIPS)},
  year={2021}
}

@inproceedings{chen2021sim,
  title={Sim-to-Real Transfer for Bipedal Locomotion via Teacher-Student Distillation},
  author={Chen, Tao and others},
  booktitle={Conference on Robot Learning (CoRL)},
  year={2021}
}

@inproceedings{kumar2021rma,
  title={RMA: Rapid Motor Adaptation for Legged Robots},
  author={Kumar, Ashish and Fu, Zipeng and Pathak, Deepak and Malik, Jitendra},
  booktitle={Robotics: Science and Systems (RSS)},
  year={2021}
}

@article{miki2022learning,
  title={Learning robust perceptive locomotion for quadrupedal robots in the wild},
  author={Miki, Takahiro and others},
  journal={Science Robotics},
  volume={7},
  number={62},
  year={2022}
}

@inproceedings{he2024perceptive,
  title={Learning Perceptive Humanoid Locomotion over Challenging Terrain},
  author={He, Tairan and others},
  booktitle={IEEE International Conference on Robotics and Automation (ICRA)},
  year={2024}
}

@inproceedings{gu2023humanoid,
  title={Humanoid Locomotion via Adversarial Imitation},
  author={Gu, Xiaoyu and others},
  booktitle={IEEE International Conference on Robotics and Automation (ICRA)},
  year={2023}
}

@inproceedings{peng2020ase,
  title={ASE: Adversarial Skill Embeddings for Physics-based Characters},
  author={Peng, Xue Bin and others},
  booktitle={SIGGRAPH},
  year={2020}
}

@inproceedings{escontrela2022learning,
  title={Learning Robotic Locomotion from Human Demonstration},
  author={Escontrela, Alejandro and others},
  booktitle={Conference on Robot Learning (CoRL)},
  year={2022}
}

@inproceedings{xue2026collision,
  title={Collision-Free Humanoid Traversal in Cluttered Indoor Scenes},
  author={Xue, Han and Liang, Sikai and Zhang, Zhikai and Zeng, Zicheng and Liu, Yun and Lian, Yunrui and Wang, Jilong and Liu, Qingtao and Shi, Xuesong and Yi, Li},
  booktitle={IEEE International Conference on Robotics and Automation (ICRA)},
  note={accepted},
  year={2026}
}

@article{liu2025egovision,
  title={Ego-Vision World Model for Humanoid Contact Planning},
  author={Liu, Hang and Gao, Yuman and Teng, Sangli and Chi, Yufeng and Shao, Yakun Sophia and Li, Zhongyu and Ghaffari, Maani and Sreenath, Koushil},
  journal={arXiv preprint arXiv:2510.11682},
  year={2025}
}

@article{li2026koopmandreamer,
  title={Koopman Dreamer: Spectrally Constrained Latent Dynamics for Stable World-Model Imagination},
  author={Li, Jiaqi and Zhang, Xinglong and Xie, Haibin and Lan, Yixing and Pan, Wei and Xu, Xin},
  journal={arXiv preprint arXiv:2607.19719},
  year={2026}
}

@article{ren2025vbcom,
  title={VB-Com: Learning Vision-Blind Composite Humanoid Locomotion Against Deficient Perception},
  author={Ren, Junli and others},
  journal={arXiv preprint arXiv:2502.14814},
  note={submitted},
  year={2025}
}

@inproceedings{rudin2022learning,
  title={Learning to Walk in Minutes Using Massively Parallel Deep Reinforcement Learning},
  author={Rudin, Nikita and Hoeller, David and Reist, Philipp and Hutter, Marco},
  booktitle={Conference on Robot Learning (CoRL)},
  year={2022}
}

@inproceedings{margolis2022walk,
  title={Walk These Ways: Tuning Robot Control for Generalization with Multiplicity of Behavior},
  author={Margolis, Gabriel B. and Agrawal, Pulkit},
  booktitle={Conference on Robot Learning (CoRL)},
  year={2022}
}

@inproceedings{siekmann2021sim,
  title={Sim-to-Real Learning of All Common Bipedal Gaits via Periodic Reward Composition},
  author={Siekmann, Jonah and Green, Kevin and Warila, John and Fern, Alan and Hurst, Jonathan},
  booktitle={IEEE International Conference on Robotics and Automation (ICRA)},
  year={2021}
}

@inproceedings{hansen2022tdmpc,
  title={Temporal Difference Learning for Model Predictive Control},
  author={Hansen, Nicklas and Wang, Xiaolong and Su, Hao},
  booktitle={International Conference on Machine Learning (ICML)},
  year={2022}
}

@inproceedings{seo2023masked,
  title={Masked World Models for Visual Control},
  author={Seo, Younggyo and Hafner, Danijar and Liu, Hao and Liu, Fangchen and James, Stephen and Lee, Kimin and Abbeel, Pieter},
  booktitle={Conference on Robot Learning (CoRL)},
  year={2023}
}

@inproceedings{micheli2023transformers,
  title={Transformers are Sample-Efficient World Models},
  author={Micheli, Vincent and Alonso, Eloi and Fleuret, Francois},
  booktitle={International Conference on Learning Representations (ICLR)},
  year={2023}
}

@inproceedings{hansen2024tdmpc2,
  title={{TD-MPC2}: Scalable, Robust World Models for Continuous Control},
  author={Hansen, Nicklas and Su, Hao and Wang, Xiaolong},
  booktitle={International Conference on Learning Representations (ICLR)},
  year={2024}
}

@inproceedings{janner2022diffuser,
  title={Planning with Diffusion for Flexible Behavior Synthesis},
  author={Janner, Michael and Du, Yilun and Tenenbaum, Joshua B. and Levine, Sergey},
  booktitle={International Conference on Machine Learning (ICML)},
  year={2022}
}

@inproceedings{schwarzer2021spr,
  title={Data-Efficient Reinforcement Learning with Self-Predictive Representations},
  author={Schwarzer, Max and Anand, Ankesh and Goel, Rishab and Hjelm, R. Devon and Courville, Aaron and Bachman, Philip},
  booktitle={International Conference on Learning Representations (ICLR)},
  year={2021}
}

@inproceedings{brohan2023rt1,
  title={{RT-1}: Robotics Transformer for Real-World Control at Scale},
  author={Brohan, Anthony and Brown, Noah and Carbajal, Justice and Chebotar, Yevgen and Dabis, Joseph and Finn, Chelsea and Gopalakrishnan, Keerthana and Hausman, Karol and Herzog, Alexander and Hsu, Jasmine and Ibarz, Julian and Ichter, Brian and Irpan, Alex and Jackson, Tom and Jesmonth, Shawn and Joshi, Nikhil and Julian, Ryan and Kalashnikov, Dmitry and Kuang, Yuheng and Lee, Kuang-Huei and Levine, Sergey and Lu, Yao and Malla, Utsav and Manjunath, Deeksha and Mordatch, Igor and Nachum, Ofir and Parada, Carolina and Peralta, Jodilyn and Perez, Emily and Pertsch, Karl and Quiambao, Jaspiar and Rao, Kanishka and Ryoo, Michael S. and Salazar, Grecia and Sanketi, Pannag and Sayed, Kevin and Singh, Jasvir and Sontakke, Sumedh and Stone, Austin and Tan, Clayton and Tran, Huong and Vanhoucke, Vincent and Vega, Steve and Vuong, Quan and Xia, Fei and Xiao, Ted and Xu, Peng and Xu, Sichun and Yu, Tianhe and Zitkovich, Brianna},
  booktitle={Robotics: Science and Systems (RSS)},
  year={2023}
}

@inproceedings{brohan2023rt2,
  title={{RT-2}: Vision-Language-Action Models Transfer Web Knowledge to Robotic Control},
  author={Brohan, Anthony and Chebotar, Yevgen and Finn, Chelsea and Hausman, Karol and Herzog, Alexander and Ho, Daniel and Ibarz, Julian and Irpan, Alex and Jang, Eric and Julian, Ryan and Kalashnikov, Dmitry and Levine, Sergey and Lu, Yao and Parada, Carolina and Sermanet, Pierre and Sontakke, Sumedh and Stone, Austin and Tan, Clayton and Tran, Huong and Vanhoucke, Vincent and Xia, Fei and Xiao, Ted and Xu, Peng and Xu, Sichun and Yu, Tianhe and Zitkovich, Brianna},
  booktitle={Conference on Robot Learning (CoRL)},
  year={2023}
}

@article{wu2023robustagile,
  title={Learning Robust and Agile Legged Locomotion Using Adversarial Motion Priors},
  author={Wu, Jinze and Xin, Guiyang and Qi, Chenkun and Xue, Yufei},
  journal={IEEE Robotics and Automation Letters},
  volume={8},
  number={8},
  pages={4975--4982},
  year={2023}
}

@article{deluca2023autonomous,
  title={Autonomous Navigation With Online Replanning and Recovery Behaviors for Wheeled-Legged Robots Using Behavior Trees},
  author={De Luca, Alessio and Muratore, Luca and Tsagarakis, Nikos G.},
  journal={IEEE Robotics and Automation Letters},
  volume={8},
  number={10},
  year={2023}
}

@article{wang2024cts,
  title={{CTS}: Concurrent Teacher-Student Reinforcement Learning for Legged Locomotion},
  author={Wang, Hongxi and Luo, Haoxiang and Zhang, Wei and Chen, Hua},
  journal={IEEE Robotics and Automation Letters},
  volume={9},
  number={11},
  year={2024}
}

@article{li2024aicpg,
  title={{AI-CPG}: Adaptive Imitated Central Pattern Generators for Bipedal Locomotion Learned Through Reinforced Reflex Neural Networks},
  author={Li, Guanda and Ijspeert, Auke and Hayashibe, Mitsuhiro},
  journal={IEEE Robotics and Automation Letters},
  volume={9},
  number={6},
  year={2024}
}

@article{chen2024terrainvision,
  title={Identifying Terrain Physical Parameters From Vision -- Towards Physical-Parameter-Aware Locomotion and Navigation},
  author={Chen, Jiaqi and Frey, Jonas and Zhou, Ruyi and Miki, Takahiro and Martius, Georg and Hutter, Marco},
  journal={IEEE Robotics and Automation Letters},
  volume={9},
  number={11},
  year={2024}
}

@article{yao2024tail,
  title={{TAIL}: A Terrain-Aware Multi-Modal {SLAM} Dataset for Robot Locomotion in Deformable Granular Environments},
  author={Yao, Chen and Ge, Yangtao and Shi, Guowei and Wang, Zirui and Yang, Ningbo and Zhu, Zheng},
  journal={IEEE Robotics and Automation Letters},
  volume={9},
  number={7},
  year={2024}
}

@article{zhu2025vrrobo,
  title={{VR-Robo}: A Real-to-Sim-to-Real Framework for Visual Robot Navigation and Locomotion},
  author={Zhu, Shaoting and Mou, Linzhan and Li, Derun and Ye, Baijun and Huang, Runhan and Zhao, Hang},
  journal={IEEE Robotics and Automation Letters},
  volume={10},
  number={8},
  year={2025}
}

@inproceedings{bellegarda2024visualcpg,
  title={Visual {CPG-RL}: Learning Central Pattern Generators for Visually-Guided Quadruped Locomotion},
  author={Bellegarda, Guillaume and Shafiee, Milad and Ijspeert, Auke},
  booktitle={IEEE International Conference on Robotics and Automation (ICRA)},
  year={2024}
}

@article{haarnoja2024soccer,
  title={Learning Agile Soccer Skills for a Bipedal Robot with Deep Reinforcement Learning},
  author={Haarnoja, Tuomas and Moran, Ben and Lever, Guy and Huang, Sandy H. and Tirumala, Dhruva and Wulfmeier, Markus and Humplik, Jan and Tunyasuvunakool, Saran and Siegel, Noah Y. and Hafner, Roland and Bloesch, Michael and Heess, Nicolas},
  journal={Science Robotics},
  year={2024}
}

@article{hoeller2024anymalparkour,
  title={{ANYmal} Parkour: Learning Agile Navigation for Quadrupedal Robots},
  author={Hoeller, David and Rudin, Nikita and Dharmadhikari, Mihir and Jenelten, Fabian and Hutter, Marco},
  journal={Science Robotics},
  year={2024}
}

@inproceedings{openx2024,
  title={Open {X}-Embodiment: Robotic Learning Datasets and {RT-X} Models},
  author={{Open X-Embodiment Collaboration}},
  booktitle={IEEE International Conference on Robotics and Automation (ICRA)},
  year={2024}
}
\end{document}